\documentclass[conference]{IEEEtran}
\IEEEoverridecommandlockouts
\usepackage{cite}
\usepackage{amsmath,amssymb,amsfonts}
\usepackage{algorithmic}
\usepackage{graphicx}
\usepackage{textcomp}
\usepackage{xcolor}
\usepackage{comment}
\usepackage{booktabs}
\graphicspath{{figures/}}
\usepackage{svg}
\usepackage{caption}
\usepackage{subcaption}
\usepackage{cite}
\usepackage{flushend}
\newcommand{\TODO}[1]{{\color{black}#1}} 
\usepackage{hyperref}

\usepackage[normalem]{ulem}

\def\BibTeX{{\rm B\kern-.05em{\sc i\kern-.025em b}\kern-.08em
    T\kern-.1667em\lower.7ex\hbox{E}\kern-.125emX}}

\usepackage{tikz}

\newcommand\copyrighttext{%
  \footnotesize \textcopyright 2023 IEEE. Personal use of this material is permitted.  Permission from IEEE must be obtained for all other uses, in any current or future media, including reprinting/republishing this material for advertising or promotional purposes, creating new collective works, for resale or redistribution to servers or lists, or reuse of any copyrighted component of this work in other works.}
\newcommand\copyrightnotice{%
\begin{tikzpicture}[remember picture,overlay]
\node[anchor=south,yshift=10pt] at (current page.south) {\fbox{\parbox{\dimexpr\textwidth-\fboxsep-\fboxrule\relax}{\copyrighttext}}};
\end{tikzpicture}%
}

\newcommand\blfootnote[1]{
	\begingroup
	\renewcommand\thefootnote{}\footnote{\vspace{-0.8cm} #1}
	\addtocounter{footnote}{-1}
	\endgroup
}

\begin{document}

\title{
Occlusion Robust 3D Human Pose Estimation with StridedPoseGraphFormer and Data Augmentation
}

\author{
 \centering
  \IEEEauthorblockN{Soubarna Banik\IEEEauthorrefmark{1}, Patricia Gschoßmann\IEEEauthorrefmark{1}, Alejandro Mendoza Garc\'{i}a\IEEEauthorrefmark{2}, Alois Knoll\IEEEauthorrefmark{1}}
  \\
  \IEEEauthorblockA{
    \begin{tabular}{cc}
        \textit{TUM School of Computation, Information and Technology}\IEEEauthorrefmark{1}
        & \textit{reFit Systems GmbH}\IEEEauthorrefmark{2}  \\
        \textit{Technical University of Munich, Munich, Germany}                        & \textit{Munich, Germany}                         \\
        \{soubarna.banik, patricia.gschossmann\}@tum.de, knoll@in.tum.de                       & alejandro@refit-systems.com
    \end{tabular}
    }
\thanks{Code is available at {\url{https://github.com/baniks/StridedPoseGraphFormer}}.}
}

\maketitle
\blfootnote{\copyrightnotice}{To appear in Proc. IEEE IJCNN 2023, June 18-23, 2023}

\begin{abstract}
Occlusion is an omnipresent challenge in 3D human pose estimation (HPE).
In spite of the large amount of research dedicated to 3D HPE, only a limited number of studies address the problem of occlusion explicitly.
To fill this gap, we propose to combine exploitation of spatio-temporal features with synthetic occlusion augmentation  during training to deal with occlusion.
To this end, we build a spatio-temporal 3D HPE model, StridedPoseGraphFormer based on graph convolution and transformers, and train it using occlusion augmentation.
Unlike the existing occlusion-aware methods, that are only tested for limited occlusion,
we extensively evaluate our method for varying degrees of occlusion. 
We show that our proposed method compares favorably with the state-of-the-art (SoA).
Our experimental results also reveal that in the absence of any occlusion handling mechanism, the performance of SoA 3D HPE methods degrades significantly when they encounter occlusion.
\end{abstract}


\section{Introduction}
3D monocular Human Pose Estimation (HPE) research has made great advances in recent years \cite{banik20213d,zhao2022graformer,zheng20213d,li2022exploiting}.
The state-of-the-art (SoA) methods solve the problem in two parts: predicting 2D joint locations from images, and lifting them to 3D poses using only the 2D poses as input.
Unlike single frame pose estimation methods, video-based HPE use both spatial and temporal information to predict temporally consistent 3D poses. 
Both single frame and spatio-temporal methods have shown remarkable accuracy on benchmark datasets. 
However, previous studies have shown that existing 3D HPE methods fail in case of occlusion and cannot recover the 3D poses correctly \cite{ghafoor2022quantification}.
Partial or complete occlusion of the body joints is common and can occur due to self-occlusion or by some external objects.
Therefore, it is important to evaluate these methods under varying degrees of occlusion.
Spatio-temporal methods perform comparatively better than single frame models.
However, they are evaluated for limited occlusion \cite{park20183d,moreno20173d}, for example for small number of joints (2-3) and only for one frame, which does not represent real-world scenarios.
Although, Ghafoor et al.~\cite{ghafoor2022quantification} report several experiments with large number of occluded joints, key details of the experimental setup are missing, such as total number of occluded frames and input type.
In absence of a reproducible and thorough occlusion robustness study, we evaluate existing single-frame \cite{banik20213d,zhao2022graformer} and multi-frame 3D HPE models \cite{li2022exploiting} for large-scale occlusion, in terms of number of occluded joints as well as number of occluded frames.
\begin{figure} [t]
     \centering
     \begin{subfigure}[]{\linewidth}
         \centering
         \includegraphics[width=\linewidth]{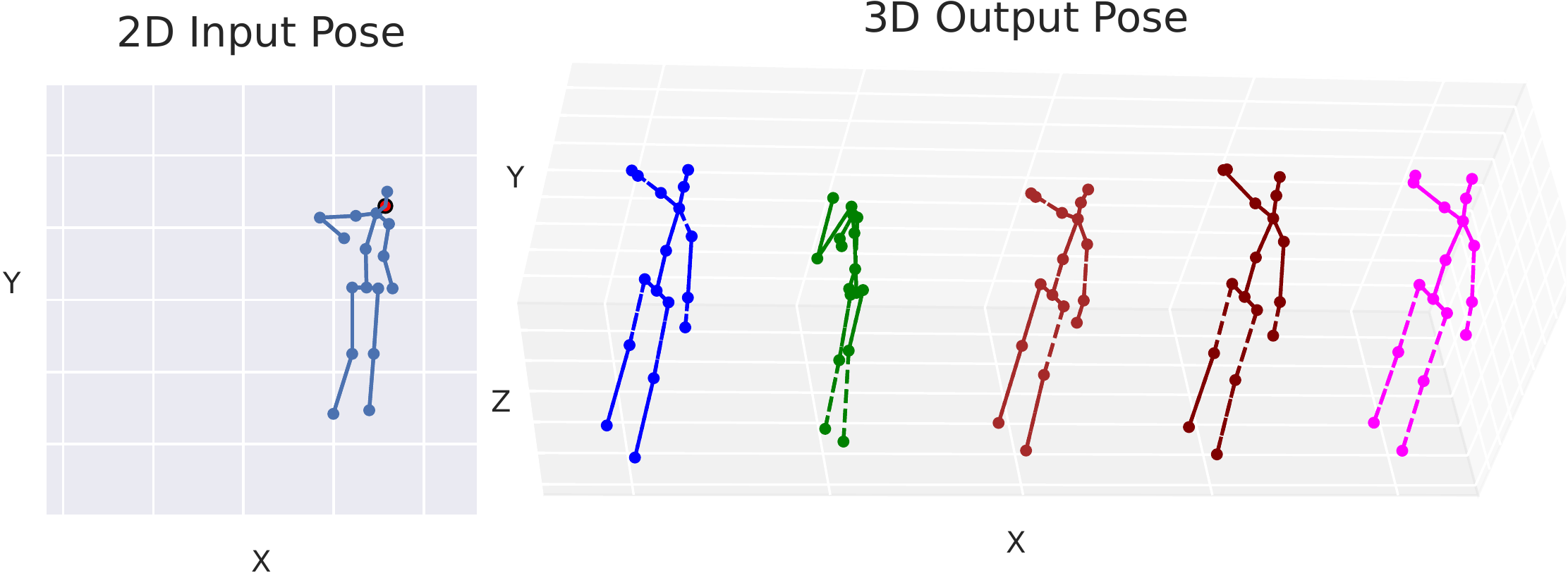}
     \end{subfigure}
     \begin{subfigure}[]{\linewidth}
         \centering
         \includegraphics[width=\linewidth]{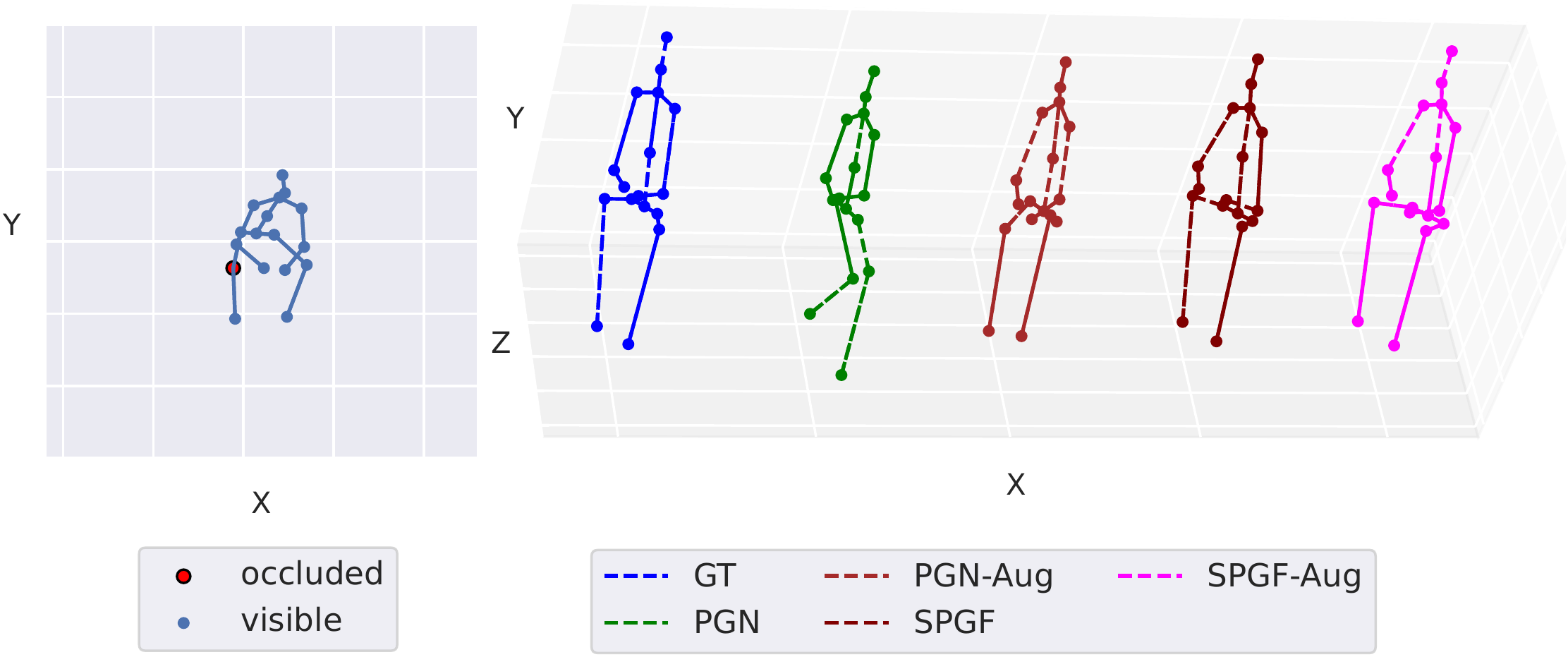}
     \end{subfigure}
    \caption{Occlusion robust 2D-to-3D pose lifting. (Left) 2D Input Pose with occluded joint highlighted in red. (Right) Ground-truth 3D pose and predictions by baseline models: PoseGraphNet (PGN) \cite{banik20213d}, PoseGraphNet-Aug (PGN-Aug), StridedPoseGraphFormer (SPGF) and StridedPoseGraphFormer-Aug (SPGF-Aug). Only central input frame is shown for visualization purpose. PGN fails to predict correct pose, whereas StridedPoseGraphFormer-Aug can recover from the occlusion.}
        \label{fig:demo}
\end{figure}

Existing occlusion-aware methods employ different techniques for emulating occlusion during training \cite{cheng2019occlusion,moreno20173d}.
However, the spatio-temporal models used in these studies do not follow the latest architectural features of this field.
With recent advances in video-based 3D HPE using graph convolution \cite{Cai2019} and transformers \cite{zheng20213d,li2022exploiting}, it is necessary to revise these studies with newer architectures.
In this paper, we propose a novel spatio-temporal 3D HPE model, \emph{StridedPoseGraphFormer} based on graph convolution and transformers. 
Although, transformer-based methods have shown exceptional performance on benchmark datasets \cite{zheng20213d,li2022exploiting}, they use multiple transformer encoders for processing both spatial and temporal features and are therefore resource intensive.
On the other hand, graph convolution-based methods \cite{banik20213d,zhao2022graformer} are light-weight and can capture structural features well.
With StridedPoseGraphFormer we aim to utilize the benefits of both genres to come up with a light-weight spatio-temporal 3D HPE method.
To improve the endurance of the model under occlusion, we augment the training dataset with synthetic occlusion, and force the model to predict correct 3D pose despite the missing keypoints in the input.
The resulting model, which is trained using occlusion augmentation is called StridedPoseGraphFormer-Aug.

We evaluate StridedPoseGraphFormer on the benchmark Human3.6M \cite{Ionescu2014} dataset.
With our tandem solution for improving occlusion robustness using (a) spatio-temporal information and (b) occlusion augmentation strategies, our model outperforms the state-of-the-art methods in adverse occlusion situations including extreme cases.
Fig.~\ref{fig:demo} shows examples of occlusion scenarios where a joint is occluded synthetically in the input by setting it to zero.
The outputs of several baseline methods are demonstrated on the right side. 
SoA 2D-to-3D lifting method PoseGraphNet (PGN) \cite{banik20213d} fails to infer the correct 3D position of the joint. Additionally, the accuracy of the adjacent joints are also affected.
Our proposed methods StridedPoseGraphFormer (SPGF) and StridedPoseGraphFormer-Aug (SPGF-Aug) can recover the position of the occluded joints  successfully. \\

\noindent Our contributions are three-fold:
\begin{itemize}
    \item We propose a light-weight spatio-temporal 3D HPE method \emph{StridedPoseGraphFormer} using graph convolution and transformers for 2D-to-3D lifting.
    \item In StridedPoseGraphFormer-Aug, we propose two solutions for ensuring occlusion robustness: exploitation of spatio-temporal relationships and synthetic occlusion during training using data augmentation.
    \item We provide an extensive occlusion robustness analysis of our proposed method as well as existing image- and video-based 3D HPE methods for large-scale occlusion, in terms of number of occluded joints and occluded frames.
\end{itemize}

\section{Related Work}
\textbf{3D Human Pose Estimation:} SoA methods for 3D human pose estimation can be broadly categorized into two groups: (a) Direct Image to 3D pose regression, and (b) 2D-to-3D lifting methods, where 3D pose is \emph{lifted} from only 2D input pose. 
Given the promising results of the latter, in recent years extensive research has been conducted in this area. The latest 2D-to-3D lifting methods can be further grouped into graph convolution-based methods and transformer-based methods. 

\emph{Graph Convolution Network (GCN)- based approaches} use a graph representation of the human body and apply graph convolution to extract structural features.
Cai et al.~\cite{Cai2019} proposed one of the first methods that use GCN for 3D HPE.
The authors specifically use a spatio-temporal GCN with U-Net architecture to capture multi-scale features.
Zhao et al.~\cite{zhao2019semantic} propose semantic graph convolution to overcome the limited receptive field of graph convolution. Their method can extract semantic information such as local and global node relationships.
Xu et al.~\cite{xu2021graph} also extract multi-scale features by using skeletal pooling and unpooling. 
Neighborhood-specific \cite{Cai2019} or joint-specific \cite{xu2021graph} individual weights have been shown to improve the results.
In contrast to the hourglass architecture of \cite{Cai2019,xu2021graph}, PoseGraphNet \cite{banik20213d} is a light-weight GCN-based 3D HPE method, that does not use any graph pooling mechanism.
The authors use adaptive adjacency matrices to capture long-range relationships between joints.
GraFormer by Zhao et al.~\cite{zhao2022graformer} uses transformer and attention for the same purpose.

\emph{Transformer-based approaches:} 
Transformers \cite{vaswani2017attention} can efficiently capture global correlations with self-attention.
Visual Transformer (ViT) \cite{vit} is one of the first applications of transformers in vision for image classification.
Recently 2D-to-3D HPE from video methods \cite{zheng20213d,li2022exploiting,li2022mhformer,shan2022p,zhang2022mixste} have also used transformers due to their ability to learn strong temporal representations over long sequence data.
Zheng et al.~\cite{zheng20213d} propose PoseFormer, a model based on ViT \cite{vit}, which uses transformers for capturing both spatial and temporal correlations. 
Li et al.\cite{li2022exploiting} use Vanilla Transformer Encoder and strided convolution 
to gradually shrink the sequence length and skip the redundant information in adjacent frames.
Unlike \cite{zheng20213d,li2022exploiting}, Shan et al. \cite{shan2022p} divide the extraction of spatial and temporal information into two stages to reduce
the complexity of this task.
MHFormer \cite{li2022mhformer} addresses the problem of ambiguity in monocular 2D-to-3D lifting methods, by proposing a three-stage approach. They produce multiple initial hypotheses for the 3D pose, which are utilized to synthesize the final prediction.
Unlike other SoA methods, MixSTE \cite{zhang2022mixste} uses its temporal transformer module to model the temporal motion of individual joints, which helps in obtaining better results.

\textbf{Occlusion-aware approaches:}
Occlusion is one of the most challenging problems in human pose estimation, but there is very little work that explicitly models the problem of occlusion, and even fewer studies perform stress testing for occlusion.
Comparatively older approaches \cite{moreno20173d,park20183d} apply different occlusion handling mechanisms for image-based static 3D HPE. 
Moreno-Noguer \cite{moreno20173d} uses a distance matrix for implementing occlusion robustness.
Park et al. \cite{park20183d} use relational dropout, an occlusion augmentation technique where a limb is masked during training.
Both approaches evaluate their methods for occlusion, but only for a small number of joints (2-3).
As their methods use only a single frame, they do not represent real-world scenarios, where joints are missing for a span of time.
Recent approaches \cite{cheng2019occlusion,cheng20203d,gu2021exploring,ghafoor2022quantification} employ different occlusion augmentation techniques for spatio-temporal data.
Cheng et al. \cite{cheng2019occlusion} use a \emph{cylinder man} model for deriving self-occlusion labels that they use for augmenting the training data. For external occlusion, they mask some joints randomly.
Cheng et al. \cite{cheng20203d} also occlude joints during training following different masking mechanisms and by introducing noise to the input.
Long-term occlusions are not applied for both approaches.
Moreover, the authors only report how occlusion augmentation helps the methods to improve the performance on standard datasets, but do not provide any occlusion-specific evaluation.
Gu et al. \cite{gu2021exploring} create a moving camera multi-person dataset that intentionally includes occlusions. 
They augment their training dataset with long-term occlusion, replicating real-world scenarios.
Their method experiences 17.7\% increase in error for 50\% occlusion of the input sequence compared to no-occlusion on Human3.6M dataset.
However, they do not emulate long-term occlusion at test time.
Ghafoor et al. \cite{ghafoor2022quantification} also implement occlusion augmentation through an occlusion guidance input. 
They evaluate their method under different occlusion scenarios, including large-scale occlusion, and outperform the state-of-the-art.
However, key experiment setup information such as training details, input type (ground truth/predicted pose), and the number of occluded frames are missing.
The authors evaluate their method for complete occlusion for a small number (3-5) of consecutive frames.
In our view, it is more important to test the performance for longer occlusion of a small number of joints, which occurs more frequently in the real world.\\

Existing occlusion-aware 3D HPE methods mainly employ data augmentation techniques to enforce occlusion during training. 
However, they lack extensive evaluation for large-scale and long-term occlusion robustness, and also comparative analysis with the SoA.
A majority of the occlusion-aware methods use spatio-temporal models, but do not quantitatively evaluate how the temporal information contributes to dealing with occlusion.
Also, the spatio-temporal models mostly use temporal convolution \cite{cheng2019occlusion,gu2021exploring,ghafoor2022quantification}, and not the latest architectural designs for spatio-temporal 3D HPE \cite{zheng20213d,li2022exploiting}.
To address these gaps, we develop a novel graph convolution- and transformer-based spatio-temporal 3D HPE model.
We evaluate our approach, as well as the latest spatio-temporal approaches, for varying degrees of occlusion and provide a detailed comparison with the SoA.

\section{Method}
We propose a Graph Convolution- and Transformer-based framework, \textit{StridedPoseGraphFormer} for occlusion-robust 3D Human Pose Estimation.
Central to this method is the differentiation between the spatial and temporal contexts through separate modules.
The model aims to overcome occlusion by modelling the spatial context through graph representation and graph convolution, and by exploiting the temporal context through self-attention.
Additionally, we employ data augmentation techniques to emulate occlusion scenarios.
Figure \ref{fig:Architecture} gives an overview of the framework.
The input to the system is a sequence of 2D joint coordinates in the image space, predicted by a state-of-the-art 2D pose estimation model, Cascaded Pyramid Network (CPN) \cite{chen2018cascaded}, which is further augmented with occlusion.
The model predicts the 3D joint positions of the target (central) frame, relative to the root (pelvis) joint in camera coordinate space.

\subsection{StridedPoseGraphFormer}
\label{sec:method:archi}
The proposed StridedPoseGraphFormer consists of three main components: (a) Spatial Graph Module (SGM), (b) Temporal Transformer Module (TTM) and (c) Strided Transformer module (STM), and some post-processing modules, as shown in Fig.~\ref{fig:Architecture}. 
We take inspiration from PoseGraphNet \cite{banik20213d} and StridedTransformer \cite{li2022exploiting} to model the spatial and the temporal modules respectively.

\subsubsection{Spatial Graph Module (SGM)}
\label{sec:SGM}
\begin{figure*}[t]
    \centering
    \includegraphics[width=\textwidth]{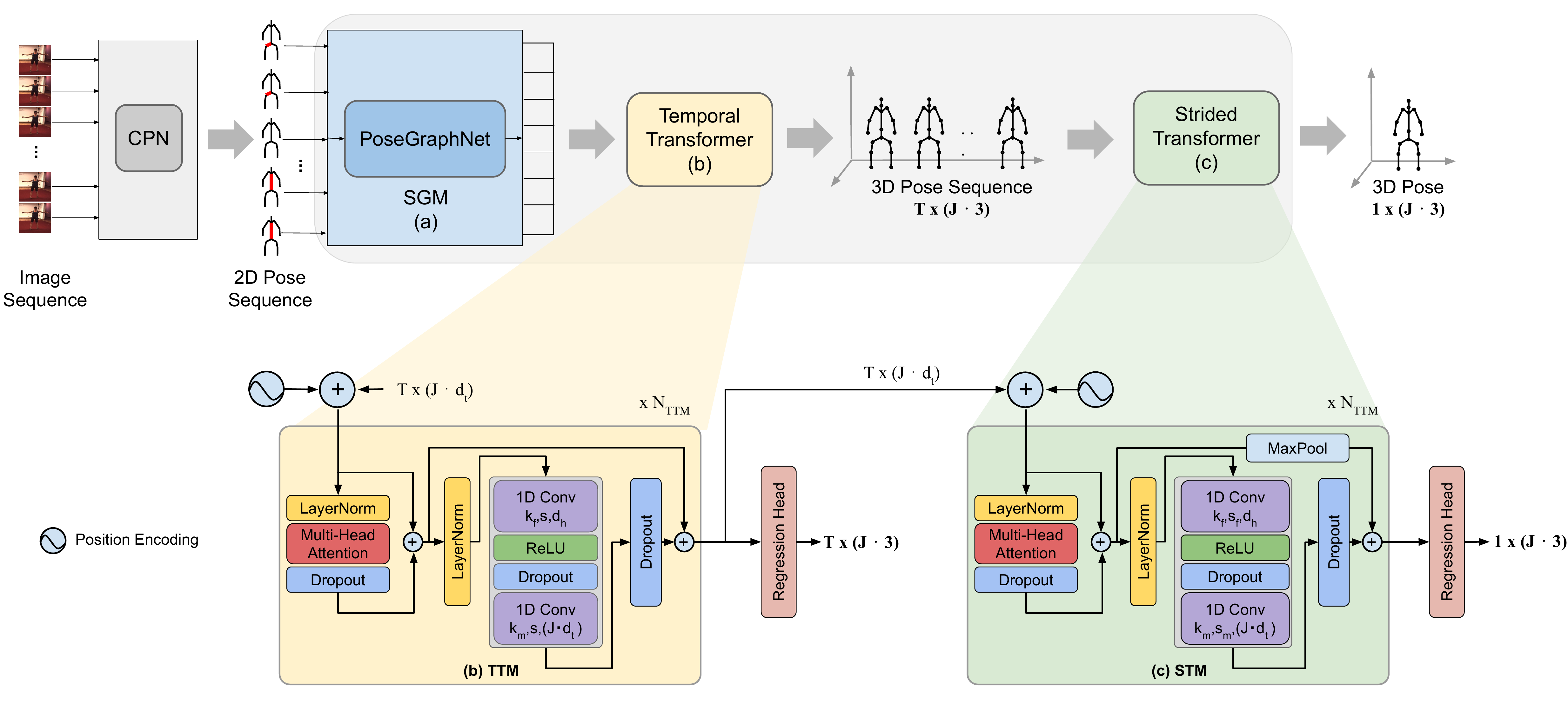}
    \caption{StridedPoseGraphFormer architecture. (Top) The overall pipeline of the framework predicting 3D joint positions of the target (central) frame from a 2D pose sequence. We use a benchmark 2D pose detector, CPN \cite{chen2018cascaded} to extract the 2D keypoints from images. We augment the 2D input pose sequence with synthetic occlusion, which are highlighted in red. The model consists of (a) Spatial Graph Module (SGM), (b) Temporal Transformer Module (TTM), and (c) Strided Temporal Module (STM). The model captures the spatial information via SGM, and the temporal context through TTM and STM to overcome occlusion. (Bottom) Expanded view of TTM and STM.}
    \label{fig:Architecture}
\end{figure*}

The first module represents the human body using a graph, 
where the nodes denote the $J$ body joints, and the edges denote the connection between the joints.
\TODO{We use} graph convolution to encode the structural information of the skeleton, which would be beneficial in case of occlusion.
\TODO{We use} a state-of-the-art GCN-based 2D-to-3D pose lifting method, PoseGraphNet \cite{banik20213d}, as SGM to model the spatial relations of the joints.
We modify PoseGraphNet by adding a graph convolution layer $gconv_h$ with $d_p$ channels before the final output layer.
The component is visualized in Fig.~\ref{fig:Architecture}(a).
See the original PoseGraphNet paper~\cite{banik20213d} for more details.
The module takes the 2D pose $x_t \in \mathbb{R}^{1 \times (J \cdot 2)}$ with $J$ joints at frame $t$ as input, and the output of $gconv_h$, 
$\tilde{P}_t \in \mathbb{R}^{J \times d_p}$ for each frame $t$ is flattened to a vector $p_t \in \mathbb{R}^{1 \times (J \cdot d_p)}$.
The vectors $\{p_0, p_1, \ldots, p_{T-1}\}$ from the $T$ input frames are then concatenated as $Z_0 \in \mathbb{R}^{T \times (J \cdot d_p)}$ and passed to the next module.

\subsubsection{Temporal Transformer Module (TTM)}
The second module is a stack of $N_{TTM}$ transformer encoders that process the entire sequence of frames to exploit the long-range temporal context and enforce temporal smoothness.
This will also help the model in case of occlusion to rely on the neighboring frames and compensate for the missing joint information in a particular frame.
This module is inspired by the Vanilla Transformer Encoder (VTE) used in \cite{li2022exploiting}.
However, we modify the encoders by replacing the standard fully-connected layers in the Feed-Forward Network (FFN) with 1D convolutions of kernel size $k_f$ and $k_m$ to better integrate local information across all frames.
The decisive factor is that a stride of $s = 1$ is used in all convolutional layers to process the entire sequence without shrinking it.
The module is built upon the flattened and concatenated outputs of the SGM and takes $Z_0$ as input.
A learnable temporal position encoding is added to the input to retain frame position information.
The resulting features are then passed to the encoders to produce the final output $Z_1 \in \mathbb{R}^{T \times (J \cdot d_p)}$.
The component is visualized in Fig.~\ref{fig:Architecture}(b).

\subsubsection{Strided Transformer Module (STM)}
The third module consists of a stack of $N_{STM}$ Strided Transformer Encoders (STEs) from~\cite{li2022exploiting} that exploit temporal features on a multi-scale basis. 
It takes $Z_1$ as input and progressively shrinks the entire sequence via two strided convolution layers with stride $s_m$.
Different learnable position encodings are applied before each encoder due to the shrinking input sequence.
The final output $z_2 \in \mathbb{R}^{1 \times (J \cdot d_p)}$ encodes the 3D pose of the center frame.
The component is visualized in Fig.~\ref{fig:Architecture}(c).
The reader is referred to~\cite{li2022exploiting} for further details about this module.

\subsubsection{Post-Processing Modules}
Both full sequence and single target frame are incorporated into the framework as in the full-to-single supervision scheme of StridedTransformer~\cite{li2022exploiting}.
To this end, two separate regression heads, each with batch normalization and a 1D convolutional layer, are applied to the outputs of the TTM and the STM, respectively.
The results $\tilde{X} \in \mathbb{R}^{T \times (J \cdot 3)}$ and $\tilde{x} \in \mathbb{R}^{1 \times (J \cdot 3)}$ correspond to the 3D pose predictions of the entire sequence and the target frame, respectively.
Finally, the pose refinement module of Cai et al.~\cite{Cai2019} is applied to $\tilde{x}$ to further enhance the estimation results and output a refined 3D pose $\tilde{x}^r$. 

\subsection{Occlusion-based Data Augmentation}
\label{sec:method:occl}
The spatio-temporal features learned by the standard StridedPoseGraphFormer will be affected in case of missing joint information.
Therefore, we propose an occlusion-based data augmentation method so that the StridedPoseGraphFormer not only relies on the standard spatio-temporal features but also learns fallback options for increased test-time occlusion robustness. 
We impose synthetic occlusion by randomly setting 2D coordinates of joints to 0.
SoA 2D pose detectors \cite{chen2018cascaded} do not predict the occluded joints as 0, but as noisy outliers, which can be converted to our input format using the joints' confidence scores additionally predicted by the 2D pose models.
The masking forces the network to rely on other relevant features to predict a correct 3D pose from an incomplete sequence of 2D poses.

Given a sequence of input frames, several frames are masked by occluding various random joints.
The total number of occluded frames $q_f$ per sequence is within $[0, q_F]$ and is determined randomly using a uniform distribution.
In real-world scenarios, joints may not be detected for longer periods due to occlusion.
Therefore, the frames are occluded in subsets consisting of $q_{cf}$ consecutive frames to mimic occlusion in reality.
First, the middle frames of each subset are determined using a uniform distribution.
Then, starting from the selected frames, the subsets are formed.
Note that this can lead to overlapping of the subsets if the drawn frames are close to each other.
$q_j$ denotes the number of joints which are occluded in a frame.
Each joint $j$ is occluded with probability $p_j$ based on a prior categorical distribution $P$.
The same joints are occluded in each frame of a subset, but the occluded joints may vary from subset to subset.

\section{Experiments}

\subsection{Dataset}
\textbf{Human3.6M} \cite{Ionescu2014} is a benchmark dataset for 3D human pose estimation. It contains 3.6 million images of 7 subjects, from 4 different viewpoints performing 15 actions such as walking, eating, etc. 
As per the standard protocol \cite{Cai2019,li2022exploiting}, we use subjects 1, 5, 6, 7, and 8 for training, and subjects 9 and 11 for evaluation. 

\subsection{Evaluation Metrics}
We evaluate the joint position prediction using the \emph{Mean Per Joint Position Error} (MPJPE) metric.
It is computed as the mean Euclidean distance of the predicted 3D joints to the ground truth joints in millimeters, following  
$E_{MPJPE}(X, \tilde{X}) = \frac{1}{J}\sum_{i=0}^{J-1} ||X(i) - \tilde{X}(i)||_2^2$, where $J$ denotes the number of joints and $X, \tilde{X} \in \mathbb{R}^{J \times 3}$ denotes the true and predicted pose, respectively.
Following previous work \cite{banik20213d,Cai2019,li2022exploiting}, we report both P1 and P2 MPJPE.

\subsection{Implementation Details}
The 2D input poses are in the image coordinate system and normalized so that the image width lies within $[-1, +1]$. 
Similar to~\cite{banik20213d, zhao2022graformer}, we keep the 3D output poses in the camera coordinate system to make the pose prediction task coherent across different camera views.
The 3D joints are zero-centered by subtracting the root joint (pelvis), as we are only interested in the relative pose of the joints with respect to the root.

The input sequence length is set to $T=81$ and contains estimated 2D keypoints from Cascaded Pyramid Networks (CPN)~\cite{chen2018cascaded}.
For the SGM, we follow the PoseGraphNet network structure defined in \cite{banik20213d}, except for the newly added $gconv_h$, for which $d_p$ is set to 16.
Unlike \cite{li2022exploiting}, we use $N_{TTM} = 1$ TTM with stride factor $s = 1$.
Our STM consists of $N_{STM} = 3$ STEs \cite{li2022exploiting} with stride factors $s_f = 1$ and $s_m = \{9, 3, 3\}$.
Both TTM and STM use kernels of size $k_f = 1$ and $k_m = 3$ and $d_h = 512$ number of hidden units in the convolutional FFN, as well as $h = 8$ attention heads.

\subsection{Training Details} 
\label{sec:Training}
The spatial module is pre-trained for frame-wise 3D pose estimation following~\cite{banik20213d} for 50 epochs.
The pre-trained layers of the modified PoseGraphNet (except the last layer) are then loaded into the SGM and frozen, while the remaining layers of StridedPoseGraphFormer are trained using transfer learning.
Next, SGM is unfrozen and finetuned jointly with the rest of the network.
Finally, the pose refinement module is added and the entire pipeline is trained in an end-to-end manner.
We use AMSGrad optimizer \cite{reddi2019convergence}, batch size of 256, and a dropout rate of 0.1.
The learning rate is initially set to 0.001 and exponentially decayed by a factor of 0.95 after each epoch and 0.5 after every fifth epoch.
During pose refinement, the initial learning rate is set to 0.0001.

\textit{Loss:}  During transfer learning and fine-tuning, StridedPoseGraphFormer is supervised at both full sequence and single target frame scale using the intermediate and final results of the MPJPE as a loss function.
The network is trained in an end-to-end manner with the total loss:

\begin{equation}
\begin{aligned}
\mathcal{L} &= \frac{1}{m} \sum_{i = 0}^{m-1} (\lambda_1\mathcal{L}_1 + \lambda_2\mathcal{L}_2)\\
\text{where } \mathcal{L}_1 &= \frac{1}{m} \sum_{i=0}^{m-1} [\lambda_1 \frac{1}{t}(\sum_{t = 0}^{T - 1} \operatorname{E}_{MPJPE}(X_i(t), \tilde{X}_i(t)) \\
\text{and } \mathcal{L}_2 &= \operatorname{E}_{MPJPE}(x_i, \tilde{x}_i)],
\end{aligned}
\label{eq:loss}
\end{equation}    

\noindent where $m$ denotes the total number of training samples, $\mathcal{L}_1$, $\mathcal{L}_2$ denote the loss functions for supervision at full sequence and single target frame scale, and $T$ denotes the number of input frames per sample.
$X_i, \tilde{X}_i \in \mathbb{R}^{T \times (J \cdot 3)}$ are the true and predicted 3D pose sequence, and $x_i, \tilde{x}_i \in \mathbb{R}^{1 \times (J \cdot 3)}$ are the true and predicted 3D pose of the target frame respectively. 
$\lambda_1$ and $\lambda_2$ are set to 1.
When applying the pose refinement, the entire pipeline is trained only with $\mathcal{L}_2$ i.e.\ $\lambda_1 =0$.

\textit{Data Augmentation: } We train the model in two ways: with and without any occlusion augmentation.
We call these two models \emph{StridedPoseGraphFormer-Aug} and \emph{StridedPoseGraphFormer} respectively.
For both cases, horizontal pose flipping is applied during training and evaluation. 
For StridedPoseGraphFormer, the model is trained on the training set of Human3.6M for a total of 19 epochs, where transfer learning amounts to 8 epochs, fine-tuning to 3 epochs, and pose refinement to 8 epochs.
StridedPoseGraphFormer-Aug, on the other hand, is trained on the occlusion augmented training set of Human3.6M following sec.~\ref{sec:method:occl}. 
Joints are occluded with equal probability, i.e. $P$ is a uniform distribution with $p_j = \frac{1}{J}$.
The number of occluded joints per frame is set to $q_j = 1$.
The subset size is set to $q_{cf} = 6$.
The maximum number of occluded frames is set to $q_F = \lfloor T / 2 \rfloor = 40$.
This model is trained for a total of 15 epochs, where transfer learning amounts to 7 epochs, fine-tuning to 3 epochs, and pose refinement to 5 epochs.

\begin{table*}[tb]
    \resizebox{\textwidth}{!}{
        \begin{tabular}{l | c c c c c c c c c c c c c c c | c }
\toprule
		\textbf{Protocol \#1} & Dir. & Disc. & Eat & Greet & Phone & Photo & Pose & Purch. & Sit & SitD. & Smoke & Wait & WalkD. & Walk & WalkT. & Avg.\\
            \midrule     
		MixSTE~\cite{zhang2022mixste} ($T = 81$) & 39.8 & 43.0 & 38.6 & 40.1 & 43.4 & 50.6 & 40.6 & 41.4 & 52.2 & 56.7 & 43.8 & 40.8 & 43.9 & 29.4 & 30.3 & \textbf{42.4} \\
		StridedTransformer~\cite{li2022exploiting} ($T = 81$) & 43.3 & 45.8 & 42.7 & 44.3 & 47.8 & 53.2 & 43.4 & 41.3 & 56.8 & 61.1 & 46.9 & 44.3 & 46.7 & 32.2 & 33.4 & 45.5 \\
		MHFormer\cite{li2022mhformer} ($T = 81$) & 41.1 & 45.2 & 41.2 & 43.1 & 45.6 & 52.7 & 42.2 & 42.5 & 54.4 & 61.3 & 45.1 & 42.8 & 46.9 & 31.4 & 33.1 & 44.6 \\
		PoseFormer~\cite{zheng20213d} ($T = 81$)& 41.5 & 44.8 & 39.8 & 42.5 & 46.5 & 51.6 & 42.1 & 42.0 & 53.3 & 60.7 & 45.5 & 43.3 & 46.1 & 31.8 & 32.2 & 44.3 \\
		P-STMO~\cite{shan2022p} ($T = 81$) & 41.7 & 44.5 & 41.0 & 42.9 & 46.0 & 51.3 & 42.8 & 41.3 & 54.9 & 61.8 & 45.1 & 42.8 & 43.8 & 30.8 & 30.7 & 44.1 \\
		CrossFormer~\cite{hassanin2022crossformer} ($T = 81$) & 40.7 & 44.1 & 40.8 & 41.5 & 45.8 & 52.8 & 41.2 & 40.8 & 55.3 & 61.9 & 44.9 & 41.8 & 44.6 & 29.2 & 31.1 & 43.7 \\ \hline
		PoseGraphNet ($T = 1$) & 46.4 & 49.5 & 48.5 & 50.8 & 55.1 & 65.5 & 50.0 & 48.2 & 61.6 & 72.5 & 52.6 & 50.3 & 54.7 & 40.8 & 43.4 & 52.7 \\
        StridedPoseGraphFormer ($T = 81$) & 44.9 & 47.3 & 42.6 & 45.7 & 47.2 & 54.2 & 45.4 & 44.1 & 57.3 & 62.7 & 47.8 & 44.2 & 48.4 & 32.9 & 33.4 & 46.5 \\
        StridedPoseGraphFormer-Aug ($T = 81$) & 43.7 & 46.4 & 42.3 & 46.1 & 47.7 & 54.6 & 44.1 & 44.1 & 57.6 & 64.8 & 47.7 & 43.7 & 47.6 & 33.0 & 33.8 & 46.5 
\\\bottomrule
	\end{tabular}
    }
    \caption{Mean Per Joint Position Error (MPJPE) in millimeter on Human3.6M test set under Protocol \#1.
    $T$ refers to the number of input frames.
    CPN \cite{chen2018cascaded} predicted inputs are used. Lower MPJPE indicates better performance.}
    \label{tab:baseline:SOA}
\end{table*}

\subsection{Performance of Base Models without Occlusion}
We first evaluate the performance of the base models StridedPoseGraphFormer and StridedPoseGraphFormer-Aug on the Human3.6M testset, without any occlusion.
We also include PoseGraphNet, which does not process any temporal information and is trained without occlusion augmentation, in this experiment.
Table \ref{tab:baseline:SOA} reports the MPJPE (P1) score of the base models, along with the SoA spatio-temporal 3D HPE models.
Our two spatio-temporal models perform significantly better than the single frame PoseGraphNet model, both achieving a P1 score of 46.5mm, and a P2 score of 37.4 and 37.5 respectively.
The temporal information helps to reduce the error by 11.8\% compared to PoseGraphNet.
The performances of the spatio-temporal models are similar. 
Although StridedPoseGraphFormer-Aug is trained on the synthetically occluded training set, its performance on the original test set is not affected.
The MPJPE score of our proposed StridedPoseGraphFormer model is 9.6\% lower than the SoA model MixSTE \cite{zhang2022mixste} (42.4mm) but with only $\frac{1}{7}$\textsuperscript{th} of the parameters and nearly 5 times throughput.

\subsection{Occlusion Robustness Analysis}
To analyze how our base models deal with different types of occlusion, we conduct several experiments. 
In section \ref{sec:occl:exp1}, we evaluate all base models on a standard occlusion test set. Based on the results we choose the best performing model and evaluate it on a variety of occlusion test sets of different complexities in section \ref{sec:occl:exp2}. 
Finally, we evaluate SoA 3D HPE methods for different levels of occlusion in section \ref{sec:occl:exp3}.

\begin{table*}[t]
\centering
\resizebox{0.85\textwidth}{!}
    {    
\begin{tabular}{c|cc|cc}
\hline
              & \multicolumn{2}{c|}{No Occl. Augmentation (A)} & \multicolumn{2}{c}{Occl. Augmentation (B)}       \\
              & PoseGraphNet        & StridedPoseGraphFormer       & PoseGraphNet-Aug & StridedPoseGraphFormer-Aug                \\
              & $T= 1$              & $T= 81$                & $T=1$       & $T=81$                        \\ \hline
H36M          & 52.7                & 46.5                 & 56.9         & 46.5 \\
H36M-Occluded & 106.7               & 81.4                 & 57.8         & 47.2 \\ \hline
$\Delta$MPJPE &54.0 & 34.9 & 0.9& \textbf{0.7} \\\hline
\end{tabular}
}
\caption{\label{tab:occl:exp1} Mean Per Joint Error (MPJPE) of baseline models based on PoseGraphNet on the Human3.6M test sets (with and without occlusion). Models are trained (A) without and (B) with occlusion augmentation.
$T = 1$ and $T=81$ denote single frame and multi-frame models respectively. $\Delta$MPJPE denotes the performance drop from the original test set to the occluded test set. Lower scores are better. Best $\Delta$MPJPE is highlighted in bold.}
\end{table*}

\begin{table*}[t]
\centering
    \resizebox{0.8\textwidth}{!}{
\begin{tabular}{c|cc|cc}
\hline
              & \multicolumn{2}{c|}{No Occl. Augmentation (A)} & \multicolumn{2}{c}{Occl. Augmentation (B)}       \\
              & GraFormer        & StridedGraFormer       & GraFormer-Aug & StridedGraFormer-Aug                \\
              & $T= 1$              & $T= 81$                & $T=1$       & $T=81$                        \\ \hline
            H36M & 54.9 & 46.0 & 58.4 & 45.3 \\
            H36M-Occluded & 103.2 & 90.3 & 55.7 & 46.0 \\\hline
$\Delta$MPJPE & 48.3& 44.3& -2.7& \textbf{0.7}\\\hline     
        \end{tabular}
    }
    \caption{\label{tab:occl:exp1:graformer} Mean Per Joint Error (MPJPE) of baseline models based on GraFormer-Small on the Human3.6M test sets (with and without occlusion). Models are trained (A) without and (B) with occlusion augmentation. $T = 1$ and $T=81$ denote single frame and multi-frame models respectively. $\Delta$MPJPE denotes the performance drop from the original test set to the occluded test set. Lower scores are better. Best $\Delta$MPJPE is highlighted in bold.}
\end{table*}

\subsubsection{Experiment 1. Performance of Base Models under occlusion}
\label{sec:occl:exp1}
The goal of this experiment is to understand whether spatio-temporal information or occlusion augmentation is the best strategy to deal with occlusion.
Altogether four models are needed for this purpose: models with and without spatio-temporal information, trained with and without occlusion augmentation.
Therefore, in addition to the three base models reported in Table \ref{tab:baseline:SOA}, we also train PoseGraphNet, which does not use any temporal information, with occlusion augmentation for this experiment.
This model is referred to as PoseGraphNet-Aug.
Except the occlusion parameter settings ($q_j = 1$, $q_f = 1$), the rest of the training details are the same as that of PoseGraphNet (see sec.~\ref{sec:Training}).
For comparing the effect of missing keypoints on the base models, a standard test set is built by masking each joint i.e. $q_j = 1$ for $q_f = 30$ frames per sequence.
Altogether 17 test runs are performed for each model.
We do not enforce any consecutive frame $q_{cf}$ constraints for this experiment.

Table \ref{tab:occl:exp1} reports the mean MPJPE (P1) scores of all test runs for the four base models.
We also report the performance on the Human3.6M original test set, and the difference in performance $\Delta$MPJPE between the two test sets.
Columns (A) and (B) show the results of the models trained without and with occlusion augmentation respectively.
On the occluded test set (H36M-Occluded), the column (B) models outperform the models in column (A) by large margins.
PoseGraphNet performs the worst on the occluded set with an MPJPE of 106.7 mm.
It is evident that the single frame input data and the 2D-to-3D lifting model on its own are not sufficient to deal with the complexity of occlusion.
In comparison, StridedPoseGraphFormer with its additional knowledge from the adjacent frames can recover better.
Its MPJPE decreases 23.7\% to 81.4mm compared to PoseGraphNet.
However, this is still almost double the error it makes on the no-occlusion test set, which indicates that temporal information alone cannot rescue from the effect of missing keypoints.
Occlusion augmentation together with the spatio-temporal information helps the StridedPoseGraphFormer-Aug model to perform the best under occlusion, achieving an MPJPE of 47.2mm and a performance drop of only 0.7mm from the no-occlusion test set. 
PoseGraphNet-Aug follows this closely by attaining a $\Delta$MPJPE of 0.9mm by only incorporating occlusion augmentation.
However, employing both spatio-temporal relationships and occlusion augmentation gives the best result.

We repeat the same experiment with another spatial graph module, GraFormer-small \cite{zhao2022graformer}, and report the scores in Table \ref{tab:occl:exp1:graformer} following the same naming convention. 
Graformer-small was changed similarly to PoseGraphNet (refer \ref{sec:SGM}) and pre-trained for 50 epochs following \cite{zhao2022graformer}.
Overall the same trend as in Table \ref{tab:occl:exp1} is observed.
However, GraFormer-Aug overfits to the occluded test set, and its performance on the original test set is affected.
Nevertheless, like StridedPoseGraphFormer-Aug in Table \ref{tab:occl:exp1}, StridedGraFormer-Aug performs the best, showing the same positive effect ($\Delta$MPJPE=0.7mm) of using both spatio-temporal relationships and data augmentation with a different design of the spatial module.


\subsubsection{Experiment 2. Varying Degrees of Occlusion}
\label{sec:occl:exp2}
This experiment analyzes the effect of increasing occlusion on the performance of the proposed model.
For this experiment, we choose the best performing base model from Table \ref{tab:occl:exp1}, i.e.\ StridedPoseGraphFormer-Aug. 
We synthesize various degrees of occlusion by modulating the number of occluded joints or the number of consecutive frames with occlusion and evaluate the model.

\begin{figure}[t]
    \centering
    \includegraphics[width=\linewidth]{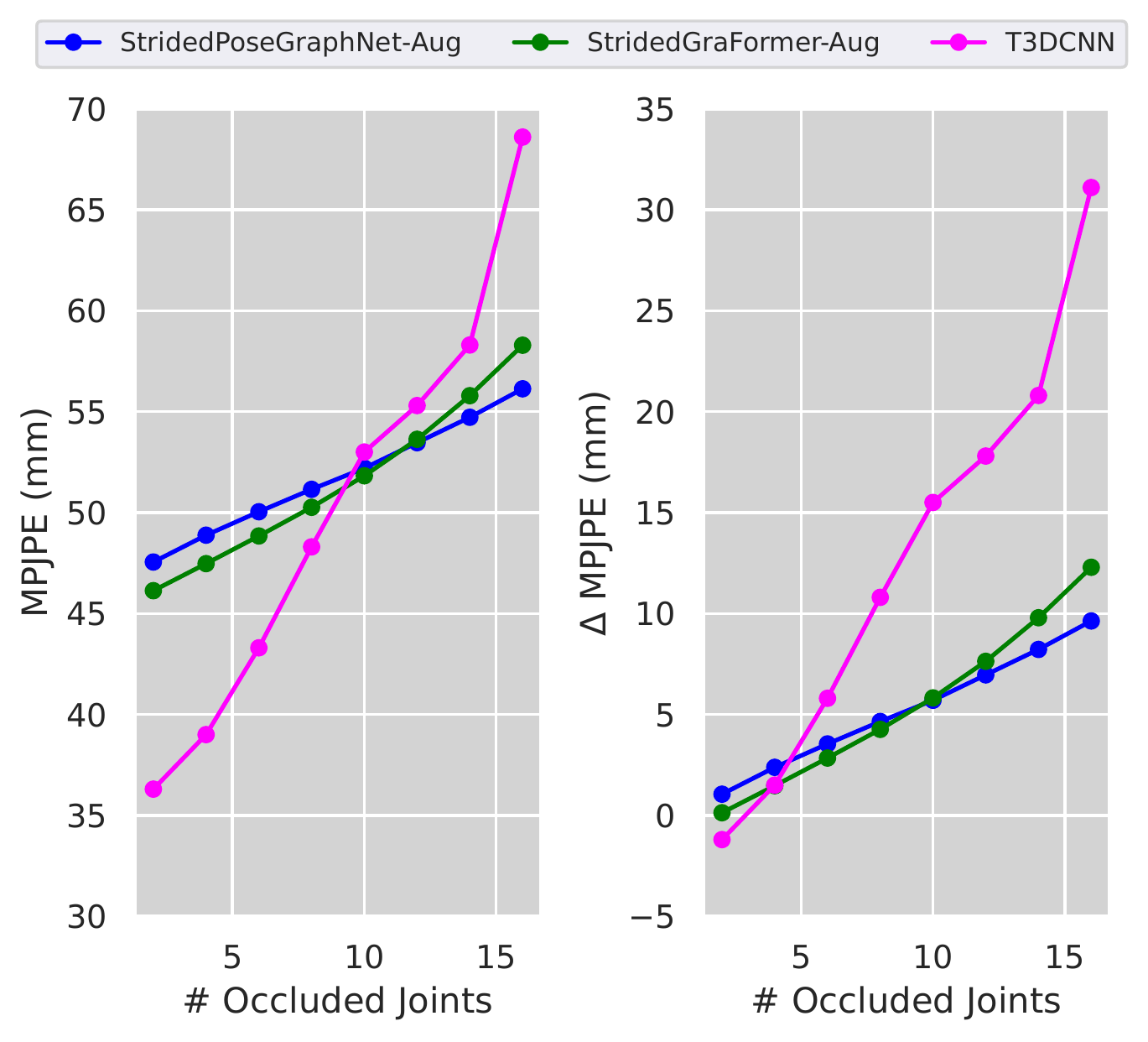}
    \caption{Performance on Human3.6M test set with increasing number of occluded joints per frame, $q_j$. (left) Mean Per Joint Position Error (MPJPE) and (right) error increase ($\Delta$MPJPE) of StridedPoseGraphFormer-Aug, StridedGraFormer-Aug (T=81) and T3DCNN (T=243) \cite{ghafoor2022quantification}. Number of consecutive occluded frames is constant $(q_{f} = 30)$.}
    \label{fig:var_occ:num_joints}
\end{figure}

\emph{Varying number of occluded joints:} Fig.~\ref{fig:var_occ:num_joints} shows the MPJPE and the error increase $\Delta$MPJPE from the corresponding no-occlusion performance, i.e.\ 46.5mm, with increasing number of occluded joints per frame. 
In each frame, $q_j=2, \dots 16$ joints are randomly occluded.
We also report the performance of StridedGraFormer-Aug and a state-of-the-art occlusion-robust 3D HPE method, T3DCNN \cite{ghafoor2022quantification} in Fig.~\ref{fig:var_occ:num_joints} for comparison.
Both StridedPoseGraphFormer and StridedGraphFormer follow our proposed occlusion robustness strategies.
Unlike our spatio-temporal models, T3DCNN uses longer sequences $T=243$ and therefore has more temporal information to counter occlusion.
Despite this advantage, the rate of error increase is the highest for this model, especially for high occlusion.
For low occlusion, T3DCNN seems to perform better than our models.
Note, the experimental setup mentioned in \cite{ghafoor2022quantification} is incomplete: they do not mention especially if ground truth 2D pose is used as input, and if occlusion is applied to all input frames.
Comparing their results for no-occlusion on Human3.6M with the latest SoA methods \cite{zheng20213d,li2022mhformer,li2022exploiting}, we deduce these are the scores for ground truth 2D input pose.
As the 2D-to-3D lifting methods work in sequence with the 2D pose detection methods, a fair evaluation is to use predicted 2D pose by benchmark 2D pose models \cite{chen2018cascaded} as input and not the ground truth 2D pose.
Given that we use CPN-predicted noisy 2D pose as input, our occlusion augmented spatio-temporal models perform considerably better.
StridedGraFormer-Aug performs better for the low range of occlusion ($q_j \in [2, 8]$), StridedPoseGraphFormer-Aug takes over for high occlusion ($q_j \in [10, 16]$).
It is to be noted, that in case of extreme occlusion with 16 out of 17 joints being occluded our proposed spatio-temporal occlusion augmented methods can recover well with an error increase of $\sim$9-12mm compared to 31.1mm by T3DCNN \cite{ghafoor2022quantification}.

\emph{Varying number of occluded frames:}  We also evaluate the performance of StridedPoseGraphFormer-Aug on increasing number of consecutive frames with occlusion.
In real applications, due to self- or external occlusion joints may not be detected for a span of time.
This kind of occlusion is more challenging, as both spatial and temporal information are missing for an interval.
Table~\ref{tab:occl:diff-degrees-frames} reports the MPJPE of StridedPoseGraphFormer-Aug for $q_{cf} \in \{1, 10, 30\}$ along with the error increase $\Delta$MPJPE from no-occlusion performance.
Only the number of consecutive frames is varied, and the number of occluded joints per frame is constant ($q_j=1$).
Although the performance drops with increasing occlusion, the error increase is limited to only 4.4mm, that too for 30 consecutively occluded frames.
\begin{table}[t]
\centering
    \resizebox{0.9\linewidth}{!}{
        \begin{tabular}{l | c c c}
            \toprule
            & $q_{cf} = 1$ & $q_{cf} = 10$ & $q_{cf} = 30$ \\
            \midrule
            MPJPE [mm] & 47.2 & 47.6 & 50.9 \\
            $\Delta$ MPJPE [mm] & 0.7 & 1.1 & 4.4 \\
            \bottomrule
        \end{tabular}
    }
    \caption{\label{tab:temp}Mean Per Joint Position Error (MPJPE) and error increase of StridedPoseGraphFormer-Aug on Human3.6M test sets with increasing number of consecutive occluded frames $q_{cf}$. Number of occluded joints per frame is constant $(q_j=1)$.}
    \label{tab:occl:diff-degrees-frames}
\end{table}

\begin{table}[t]
\centering
    \resizebox{\linewidth}{!}{
\begin{tabular}{l|c|cc}
\toprule
                   & H36M & \multicolumn{2}{c}{H36M-Occluded} \\
                   &      & $q_f = 30$       & $q_f=81$       \\ \hline
PoseFormer \cite{zheng20213d}        & 44.3 & 60.7             & 129.4         \\                   
MHFormer \cite{li2022mhformer}       & 44.6 & 68.5             & 177.2          \\
StridedTransformer \cite{li2022exploiting} & 45.5 & 78.8             & 182.5          \\
StridedPoseGraphFormer (Ours) & 46.5 & 81.4 & 144.6\\\hline
StridedPoseGraphFormer-Aug (Ours) & 46.5 & \textbf{47.2} & \textbf{56.3}\\ \hline
\end{tabular}
}
\caption{\label{tab:occl:exp3}MPJPE performance of state-of-the-art spatio-temporal 3D HPE methods with $T=81$ input frames for increasing number of occluded frames, $q_f \in \{30, 81\}$. Number of occluded joints per frame is constant $(q_j=1)$.}
\end{table}

\subsubsection{Experiment 3: Evaluation of SoA Methods under occlusion}
\label{sec:occl:exp3} 
We evaluate SoA spatio-temporal methods for 3D HPE on Human3.6M test set with different levels of occlusion.
For this experiment we choose the best performing SoA models \cite{zheng20213d,li2022mhformer,li2022exploiting} with $T=81$ input frames.
Table~\ref{tab:occl:exp3} reports the MPJPE (P1) score of the SoA methods for $q_f \in \{30,81\}$ randomly occluded frames.
The number of joints occluded per frame is constant i.e.\ $q_j=1$. 
All the spatio-temporal methods suffer significantly under occlusion, and the error increases by large margins.
This reaffirms our claim in Exp.1 in section \ref{sec:occl:exp1} that spatio-temporal information alone cannot recover from occlusion.
In contrast, the performance of the spatio-temporal occlusion augmented model StridedPoseGraphFormer-Aug is exceptional, also for the extreme occlusion case where a random joint is missing in all input frames.

In summary, our results indicate that occlusion augmentation does not affect the 3D HPE models' performance in no-occlusion test scenarios, and helps them immensely to recover from occlusion.
The best results are achieved by spatio-temporal model with occlusion augmentation, StridedPoseGraphFormer-Aug.
Compared to SoA method \cite{ghafoor2022quantification}, our proposed model reacts well to large-scale occlusion with more occluded joints and frames than it is exposed to during training.

\section{Conclusion}
We present two strategies to overcome occlusion related challenges for 3D human pose estimation: exploiting spatio-temporal information and training with occlusion augmentation.
For evaluating the solutions, we develop a novel graph convolution- and transformer-based 3D HPE model: StridedPoseGraphFormer.
We evaluate different baselines following our proposed strategies on the standard Human3.6M dataset.
Our results show that StridedPoseGraphFormer-Aug which incorporates both solutions gives the best result.
We also analyze several state-of-the-art single-frame and spatio-temporal 3D HPE models' performance under extensive occlusion, and observe that despite their top scores on no-occlusion scenarios, the performances drop significantly in occluded test sets, indicating the need for occlusion augmentation during training.
Our results show that occlusion augmentation does not affect the results on the no-occlusion test case.
Compared to the occlusion-trained SoA method \cite{ghafoor2022quantification}, our proposed model recovers the 3D pose more accurately in case of large-scale occlusion.


\bibliographystyle{IEEEtran}
\bibliography{main.bib}

\end{document}